\pdfoutput=1
\documentclass[11pt]{article}

\usepackage{ACL2023}

\usepackage{times}
\usepackage{latexsym}
\usepackage{color, colortbl, xcolor}
\usepackage{multirow}
\usepackage{mathabx}
\usepackage{subcaption}
\usepackage{graphicx}
\usepackage{tabularx}
\usepackage{booktabs}
\usepackage{stfloats}

\graphicspath{{figs/}}

\usepackage[T1]{fontenc}
\usepackage[utf8]{inputenc}

\usepackage{microtype}

\usepackage{inconsolata}

\title{Towards Adaptable and Interactive Image Captioning with Data Augmentation and Episodic Memory}

\author{Aliki Anagnostopoulou$^{1,2}$ \ \ \ \ \ Mareike Hartmann$^3$ \ \ \ \ \ Daniel Sonntag$^{1,2}$ \\
$^1$German Research Center for Artificial Intelligence (DFKI), Germany\\
$^2$Applied Artificial Intelligence (AAI), Oldenburg University, Germany \\
$^3$Department of Language Science and Technology, Saarland University, Germany \\
\texttt{\{firstname.lastname\}@dfki.de}}

\begin{document}

\maketitle

\begin{abstract}
    Interactive machine learning (IML) is a beneficial learning paradigm in cases of limited data availability, as human feedback is incrementally integrated into the training process. In this paper, we present an IML pipeline for image captioning which allows us to incrementally adapt a pre-trained image captioning model to a new data distribution based on user input.  In order to incorporate user input into the model, we explore the use of a combination of simple data augmentation methods to obtain larger data batches for each newly annotated data instance and implement continual learning methods to prevent catastrophic forgetting from repeated updates. For our experiments, we split a domain-specific image captioning dataset, namely VizWiz, into non-overlapping parts to simulate an incremental input flow for continually adapting the model to new data. We find that, while data augmentation worsens results, even when relatively small amounts of data are available, episodic memory is an effective strategy to retain knowledge from previously seen clusters.
\end{abstract}

% Kombination einfacher Methoden für die inkrementelle Verbesserung eines Image-Captioning-Systems

\section{Introduction}

\begin{figure*}[ht]
\centering
  \includegraphics[width=\linewidth]{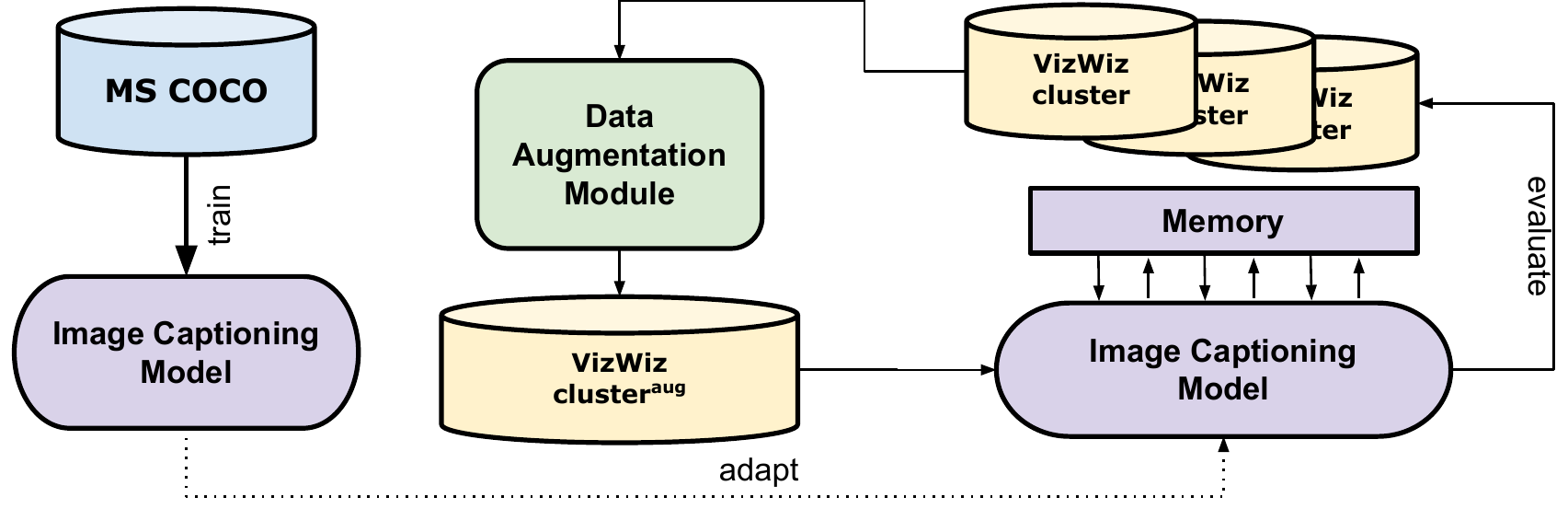}
  \caption{Our pipeline. Following the pre-training/fine-tuning paradigm, we first train our IC model on the MS COCO dataset. We then continue to train our model incrementally, by adding a new cluster each time from the VizWiz dataset, after applying DA methods on it to obtain more training data. During training on the VizWiz data for each cluster, an episodic memory module is activated, which is used to retrieve old data points from previously seen clusters.}
  \label{fig:pipeline}
\end{figure*}

% From abstract (minor changes).
Image Captioning (IC) is the task of generating a description in natural language for a given image \cite{stefanini-etal}.
For the training of most state-of-the-art IC models, large amounts of annotated training data are required \cite{zhou2019vlp}. However, whenever models need to caption user-specific images without large-scale annotations, this is an impractical requirement.
In this case, a potential solution can be found in an \textit{interactive} framework, in which the model can be efficiently adapted to new data based on user feedback \cite{ling2017teaching,shen2019learning}. Additionally, interactivity renders AI/ML-systems more user-friendly and trustworthy \cite{7349687,guo2022trust}.
% missing: \cite{7349687}

% Addition: why data augmentation? why continual learning?
In interactive ML settings, training takes place with small amounts of data, and often in an incremental manner. These properties can lead to \textit{overfitting}, on the one hand, which is the lack of generalization ability of the model, and \textit{catastrophic forgetting}, on the other hand, which refers to the drop in performance on older tasks, when a model is trained on new data. 
For our interactive approach, we tackle these problems using a combination of methods previously proposed in the literature. %Since overfitting can be tackled by generating additional datapoints, we apply data augmentation to each instance of user feedback \cite{wang-etal-2021-putting}.
To tackle overfitting, we apply data augmentation to each instance of user feedback to obtain larger batches of data, which the model is then updated on \cite{wang-etal-2021-putting}. Nevertheless, we find that this strategy fails to improve results in our image captioning task, indicating that the data augmentation methods we used are not suitable for this kind of task.
In order to prevent catastrophic forgetting, we rely on continual learning methods. In the following, we present and test an IC pipeline that can be used in an interactive setting. Our work is guided by the following research questions:

\begin{enumerate}
    \item How does data augmentation benefit a system which is trained incrementally with (simulated) user feedback? How does this system perform in few-shot scenarios?
    \item How effective is an episodic memory replay module \cite{massondautume-etal-2019-episodic} for knowledge retention from previous trainings?
\end{enumerate}

Our contributions are as follows:
\begin{itemize}
    \item We propose a lightweight continual learning IC pipeline that leverages data augmentation, which can be used in an interactive machine learning setting.
    \item We adapt a continual learning method, namely sparse memory replay, proposed by \citet{massondautume-etal-2019-episodic}, for IC.
    \item We test a combination of data augmentation methods for interactive IC in both image and text modalities.
    \item Since we report negative results for the system using data augmentation methods on the user feedback, we additionally investigate why these methods do not work in our case, and we offer some possible explanations for the deteriorating performance.
    \item We propose a method based on nominal phrase similarity between captions of different images for splitting a dataset into different tasks suitable for evaluating task-incremental continual learning when only image captions are given.
\end{itemize}

For our simulated user feedback, we use a domain-specific dataset, namely VizWiz \cite{vizwiz1,vizwiz2}, which consists of images taken by visually impaired people. We choose this dataset exactly because of this property: the quality of the images is lower than in most general-use IC datasets, resembling the image quality of user images.

\section{Related work}\label{sec:relwork}

\paragraph{Image captioning (IC)} Deep-learning based IC models \cite{xu2015show,anderson2018bottom} traditionally consist of two parts: an \textit{encoder} and a \textit{decoder}. The visual encoder breaks the image down into features or creates an intermediate representation. The decoder is a language model, which takes the encoder output as input and generates a caption. For \textit{grounded} approaches, more supervision is required: image features, such as regions, are additionally inserted into the visual encoder \cite{lu2018neural}. Following the trend in other deep learning tasks, recent approaches include large-scale vision-language pre-training, as well as generalized models that work for a variety of computer vision and vision-language tasks, including image retrieval, referring segmentation, and visual question answering \cite{zou2022xdecoder,li2022mplug}.

\paragraph{Interactive IC} Interactive IC has not gained as much attention as other ML tasks. \citet{10.1145/3372278.3390697} involve the human-in-the-loop by providing incomplete sequences as input, in addition to each image, during inference time. \citet{biswas2020XIML} extend the Show, Attend, and Tell architecture by combining high-level and low-level features, which provide explainability, as well as beam search during decoding time. 

\paragraph{Data augmentation} Data augmentation (DA) is widely applied to multiple tasks which include learning from large data, whenever there is a lack of annotated instances. It can additionally be used as a regularization technique to avoid overfitting by introducing noise into the dataset. In Computer Vision, transformations like cropping, warping, and horizontal/vertical flipping are often applied \cite{takahashi2019data,katiyar2021image}. 
% Missing here: \cite{wang2018image}

For text, augmentation methods need to be more elaborate, since image-inspired techniques often change the semantics of the text drastically. Popular methods include, but are not restricted to, EDA \cite{EDA} (including random insertion, deletion, word swap), back-translation  \cite{sennrich-etal-2016-improving,9659834}, synonym replacement and contextual augmentation \cite{kobayashi-2018-contextual,atliha2020text}, often using a pre-trained language model \cite{devlin-etal-2019-bert}. For both modalities, retrieval-based augmentation from additional resources is possible as well \cite{li2021similar}.
% Missing here: \cite{mccarthy-navigli-2007-semeval}

\paragraph{Continual Learning} In cases where a model is trained repeatedly on new data, \textit{catastrophic forgetting} \cite{kirkpatrick2017overcoming} can be observed. This refers to the degradation of model performance on older tasks when it is trained on new ones. In order to overcome this, continual learning methods are often applied. Methods such as weight regularization, encoder/decoder freezing, pseudo-labeling, and knowledge distillation, have been previously applied to IC models \cite{nguyen2019contcap,del2020ratt}. In the natural language processing domain, \citet{massondautume-etal-2019-episodic} use a combination of episodic memory replay during training and local adaptation of the model during inference. 

\section{Method}\label{sec:method}

% I am not sure if this is a nice name for this section...
In this section, we describe the approach followed, including our benchmark strategy, our DA methods, as well as the episodic memory module. Our pipeline is illustrated in \autoref{fig:pipeline}.

\subsection{Interactive IC pipeline}
\label{sec:iic}

\paragraph{Architecture} We experiment with a concrete implementation of the interactive approach outlined in \citet{hartmann2022IMLIC}. We use a PyTorch implementation of \textit{Show, Attend and Tell} \cite{xu2015show}. This architecture consists of a convolutional neural network (CNN) encoder, which is used to extract feature vectors from images, and a long-short-term memory (LSTM) decoder, which generates a caption conditioned on these vectors, with the use of attention. Following \citet{vizwizlessons}, we replace the ResNet encoder with a ResNext network \cite{Xie2016AggregatedRT}.

\begin{figure}[ht!]
    \centering
    \includegraphics[width=\linewidth]{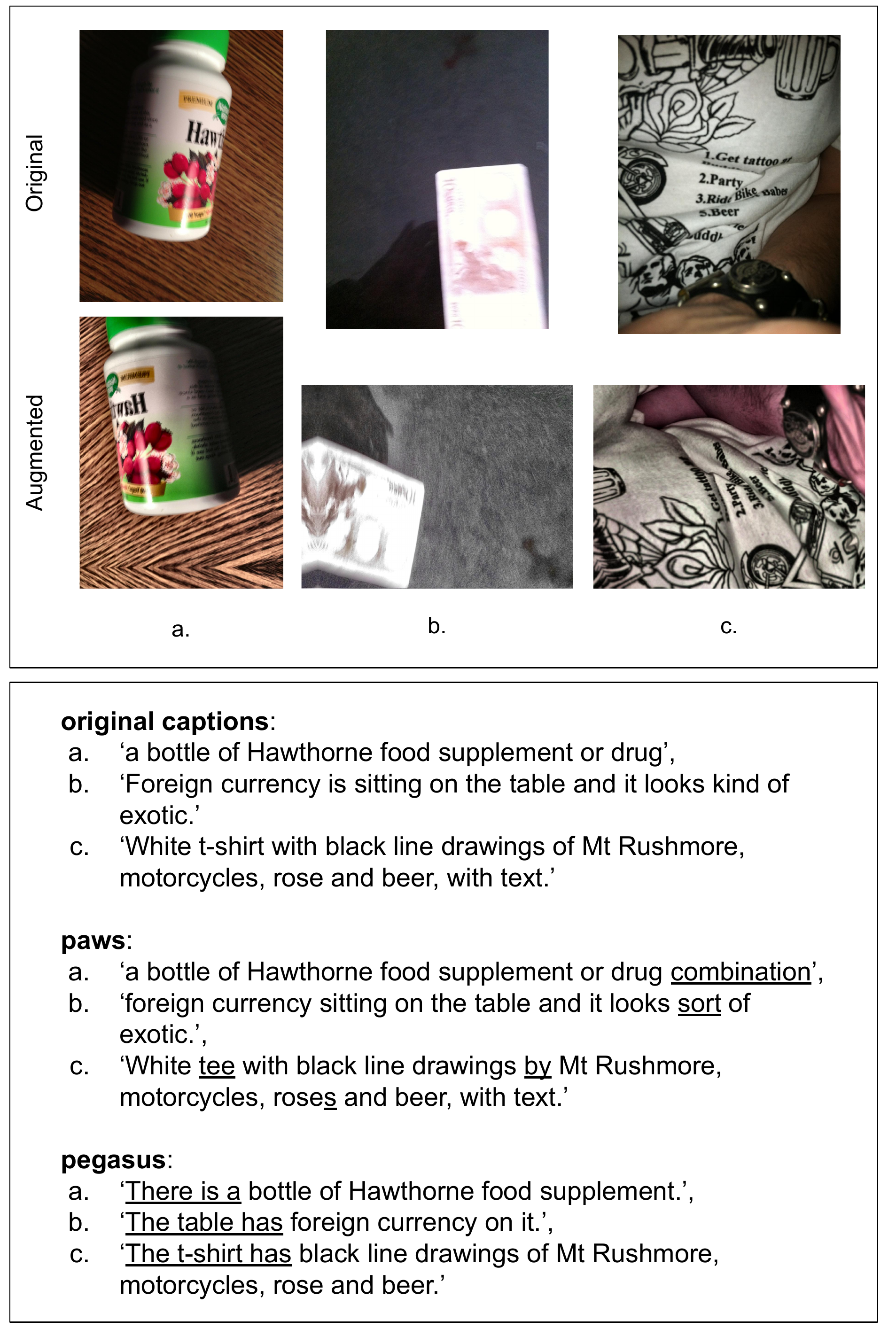}
    \caption{Generated data points generated based on the DA methods described in \autoref{sec:iic}. Top: image DA (combination of several DA methods). Bottom: text DA.}
    \label{fig:augmentation}
\end{figure}

For the decoder, an LSTM network is used. A problem arising from incremental training here is the expansion of the vocabulary. In order to tackle this problem, 
we rely on the subword vocabulary given by the BERT \cite{devlin-etal-2019-bert} tokenizer provided by Huggingface\footnote{We use \texttt{bert-base-uncased}.}.
By using a pre-trained subword tokenizer, we account for new words learned incrementally, without the need to expand the model size. The training strategy used is cross-entropy loss. 

While current state-of-the-art architectures achieve better scores, we adapt this particular architecture because of its simplicity, and because its inputs are raw images, as opposed to image features like bounding boxes and labels from object recognition models, which further decreases preprocessing time. The pipeline can potentially be adapted to any IC model that takes images as input, rather than image regions and classes. 

\paragraph{Data augmentation methods} For our experiments, we use DA on Image (\textsc{img}), Text (\textsc{txt}), and both modalities simultaneously (\textsc{both}). For \textsc{img}, we use the Albumentations \cite{info11020125} library. We create a pipeline of different operations, including CLAHE, optical and grid distortion, blur, flip, and rotation. Our goal here is to introduce noise to the input data, in order to help the model generalize better to unseen data. 
%Captions
For the \textsc{txt} modality, we aim at generating meaningful captions. For this reason, we employ two paraphrasing models provided by Huggingface, namely \texttt{pegasus\_paraphrase}, a PEGASUS \cite{zhang2019pegasus} model fine-tuned for paraphrasing, and \texttt{paws\_paraphrase}, a T5 \cite{2020t5} model trained on the PAWS \cite{paws2019naacl,pawsx2019emnlp} dataset. The reason we use two different paraphrasing tools is that we found out that the quality of the generated samples is different. In addition, paraphrasing quality drops in each tool when the number of paraphrases increases. In order to introduce more variety without compromising the quality, we decide to utilize two paraphrasing tools. In the case of combined (\textsc{both}) DA, \textsc{img} augmented images are combined with synthetically generated captions.  In every case, we generate batches that are 10 times bigger than the initial ones. Examples of generated data points can be found in \autoref{fig:augmentation}.

\paragraph{Episodic memory for lifelong learning} In order to help the model retain old knowledge when being adapted to new data, we implement a continual learning method, more specifically a sparse memory replay that operates during training. We adapt the method described by \citet{massondautume-etal-2019-episodic}: During training, some samples/experiences are written into the memory. Every training sample has a certain probability to be selected for memory writing. These experiences are then sparsely replayed (i.e. 1 sample from memory for every 200 new data points, see \autoref{sec:proc}) while the model is trained on new data. This way, the model retains information from previous training iterations with very low additional computational effort.

\begin{table}[]
\centering
\begin{tabular}{l|rrr|r|r}
\toprule
 & train & val & test & all & WT \\
 \midrule
1 & 3,332 & 954 & 2,476 & 6,762 & 10,047 \\
2 & 1,535 & 302 & 488 & 2,325 &  4,988 \\
3 & 5,668 & 1,402 & 2,199 & 9,269 & 13,497 \\
4 & 333 & 83 & 113 & 529 & 2,931\\
5 & 6,160 & 1,516 & 2,474 & 10,150 & 12,407 \\
\midrule
all & 17,028 & 4,257 & 7,750 & 29,035 & 21,955 \\
\bottomrule
\end{tabular}
\caption{VizWiz cluster (task) statistics after filtering out bad quality images (according to the procedure mentioned in \autoref{sec:datasets}). WT stands for word types.}
\label{tab:stat}
\end{table}

\subsection{Procedure and training details}\label{sec:proc}

We follow the pre-training/fine-tuning paradigm, where we first train the model on a \textit{supervised pre-training} task using a large, generic dataset, namely MS COCO \cite{lin2014microsoft} (details below). During (supervised) pre-training, we do not use any DA or continual learning method. After obtaining the best model, we continue with our incremental \textit{model adaptation}, during which we apply DA and continual learning. 

\begin{table*}[]
\centering
\resizebox{\linewidth}{!}{
\begin{tabular}{l|rrrr|rrrr|rrrr|rrrr|rrrr}
\toprule
 & \multicolumn{4}{c}{+ cluster 1 [3332]} & \multicolumn{4}{c}{+ cluster 2 [1535]} & \multicolumn{4}{c}{+ cluster 3 [5668]} & \multicolumn{4}{c}{+ cluster 4 [333]} & \multicolumn{4}{c}{+ cluster 5 [6160]} \\
 % & \multicolumn{4}{c}{3332} & \multicolumn{4}{c}{1535} & \multicolumn{4}{c}{5668} & \multicolumn{4}{c}{333} & \multicolumn{4}{c}{6160} \\
DA & \multicolumn{1}{c}{\textsc{no}} & \multicolumn{1}{c}{\textsc{img}} & \multicolumn{1}{c}{\textsc{txt}} & \multicolumn{1}{c}{\textsc{both}} & \multicolumn{1}{c}{\textsc{no}} & \multicolumn{1}{c}{\textsc{img}} & \multicolumn{1}{c}{\textsc{txt}} & \multicolumn{1}{c}{\textsc{both}} & \multicolumn{1}{c}{\textsc{no}} & \multicolumn{1}{c}{\textsc{img}} & \multicolumn{1}{c}{\textsc{txt}} & \multicolumn{1}{c}{\textsc{both}} & \multicolumn{1}{c}{\textsc{no}} & \multicolumn{1}{c}{\textsc{img}} & \multicolumn{1}{c}{\textsc{txt}} & \multicolumn{1}{c}{\textsc{both}} & \multicolumn{1}{c}{\textsc{no}} & \multicolumn{1}{c}{\textsc{img}} & \multicolumn{1}{c}{\textsc{txt}} & \multicolumn{1}{c}{\textsc{both}} \\ \midrule
1 & 18.8 & 6.4 & 15.8 & 15.3 & 12.4 & 2.2 & 11.3 & 4.4 & 15.9 & 2.4 & 13.0 & 7.3 & 12.7 & 1.9 & 9.8 & 3.9 & 11.8 & 2.8 & 9.7 & 7.1 \\
2 &  &  &  &  & 26.0 & 6.9 & 19.8 & 16.4 & 25.0 & 5.5 & 18.7 & 11.3 & 18.7 & 4.6 & 13.0 & 7.2 & 22.6 & 3.5 & 14.9 & 13.8 \\
3 & &  &  &  &  & &  &  & 27.7 & 4.2 & 24.5 & 16.3 & 21.1 & 2.3 & 16.4 & 4.9 & 22.4 & 2.9 & 16.9 & 11.9 \\
4 &  &  &  & & &  &  &  &  & & &  & 26.7 & 4.6 & 20.5 & 13.1 & 20.4 & 3.4 & 15.4 & 10.6 \\
5 &  & & & & &  &  &  &  &  &  &  &  &  &  &  & 25.9 & 3.7 & 19.2 & 15.3 \\
\midrule
all & 18.8 & 6.4 & 15.8 & 15.3 & 16.4 & 3.4 & 14.6 & 7.4 & 23.6 & 3.6 & 19.9 & 12.2 & 18.4 & 2.4 & 14.2 & 5.0 & 21.2 & 3.3 & 16.2 & 12.1 \\
\bottomrule
\end{tabular}
}
\caption{CIDEr results on our experiments on VizWiz data clustered according to the procedure described in \autoref{sec:datasets}. We start with the model resulting from the supervised pre-training step on MS COCO and continue to train this model incrementally on the VizWiz clusters (+cluster ...). We include the amount of (original) training data in brackets. DA: Data augmentation, \textsc{no}: no DA, \textsc{img}: image DA, \textsc{txt}: text DA, \textsc{both}: image and text DA. The numbers in the left column stand for clusters evaluated on. 'all' refers to the micro avg.}
\label{tab:results2}
\end{table*}

\paragraph{Training details} For the supervised pre-training step, we train our model on MS COCO in two stages: during the first training, we freeze the encoder and only train the decoder. The encoder is then trained in the second stage. For the adaptation step, we train our models on each task once. 

We train with a batch size of 32 and a learning rate of 4e-4 for the decoder. For our memory module, the replay frequency is 200, as mentioned in \autoref{sec:iic}; that means that for every 200 batches, one batch is drawn from the memory and added to the current training batch. The memory writing probability is 20\%.

We use early stopping. During our initial experiments, we trained with higher (p=10) and lower (p=2) patience values for early stopping. During our initial experiments, lower patience seems to produce better results, hence we adopt this value for our adaptation training. During supervised pre-training, we used 20 as the default patience value.

\subsection{Datasets}\label{sec:datasets}

\paragraph{Supervised pre-training step} We first train our model on the MS COCO dataset \cite{lin2014microsoft}. It contains 328k images, and it is broadly used as a pre-training dataset for vision tasks, including object recognition, object segmentation, and IC. We use the 2014 release, which contains 82,783 training and 40,504 validation images. Each image is annotated with five captions, describing the content of each image. We make use of the Karpathy splits \cite{karpathy}.

\paragraph{Adaptation} After obtaining the best possible captioning model trained on MS COCO, we train our model incrementally using VizWiz \cite{vizwiz1,vizwiz2}, a dataset consisting of images taken by visually impaired people. Since there are no test captions available, we use the validation set as our test set. A part of the training samples is used as our validation set. % , and use a part of the training samples as our validation set.

\paragraph{Dataset processing}

We want to simulate a continual learning setting where we incrementally adapt the IC model to new sets of user-specific data. For this, we split VizWiz into non-overlapping clusters representing user-specific datasets. We follow the procedure for other continual learning datasets, where data is split according to classes/concepts, and each new class/concept represents a new task \cite{del2020ratt}. As the VizWiz dataset does not contain object annotations for its images, we resort to splitting the data according to the objects mentioned in the captions, using the procedure described below. The resulting clusters resemble the user-specific data we might expect to receive from different users in a real-world setup: Whereas one user might be more interested in captioning screenshots or images of IT-related concepts, another user might be interested in captioning images of containers of food and drinks, etc. Example NPs for each cluster can be found in \autoref{app:sec:nps}.

We follow the steps below:

\begin{enumerate}
    \item We collect all nominal phrases (NPs) in the entire caption corpus. We use TextBlob\footnote{\url{https://textblob.readthedocs.io/en/dev/}} for the extraction of the NPs.
    \item From all the NPs, we choose so-called \textit{keywords}, namely phrases that appear at least 15 times in the dataset.
    \item Using GloVe \cite{pennington2014glove} embeddings, we extract word embeddings for each keyword. In case a keyword is phrasal, we average between individual word embeddings.
    \item We cluster the keyword embeddings in 5 clusters, using K-means clustering \cite{Hartigan1979}.
    \item We iterate over all captions for each image, looking for relevant keywords, and assigning them to clusters. In case one image corresponds to more than one cluster according to its keywords, we favor the smaller cluster.
\end{enumerate}

\begin{figure*}[ht]
\centering
  \includegraphics[width=\linewidth]{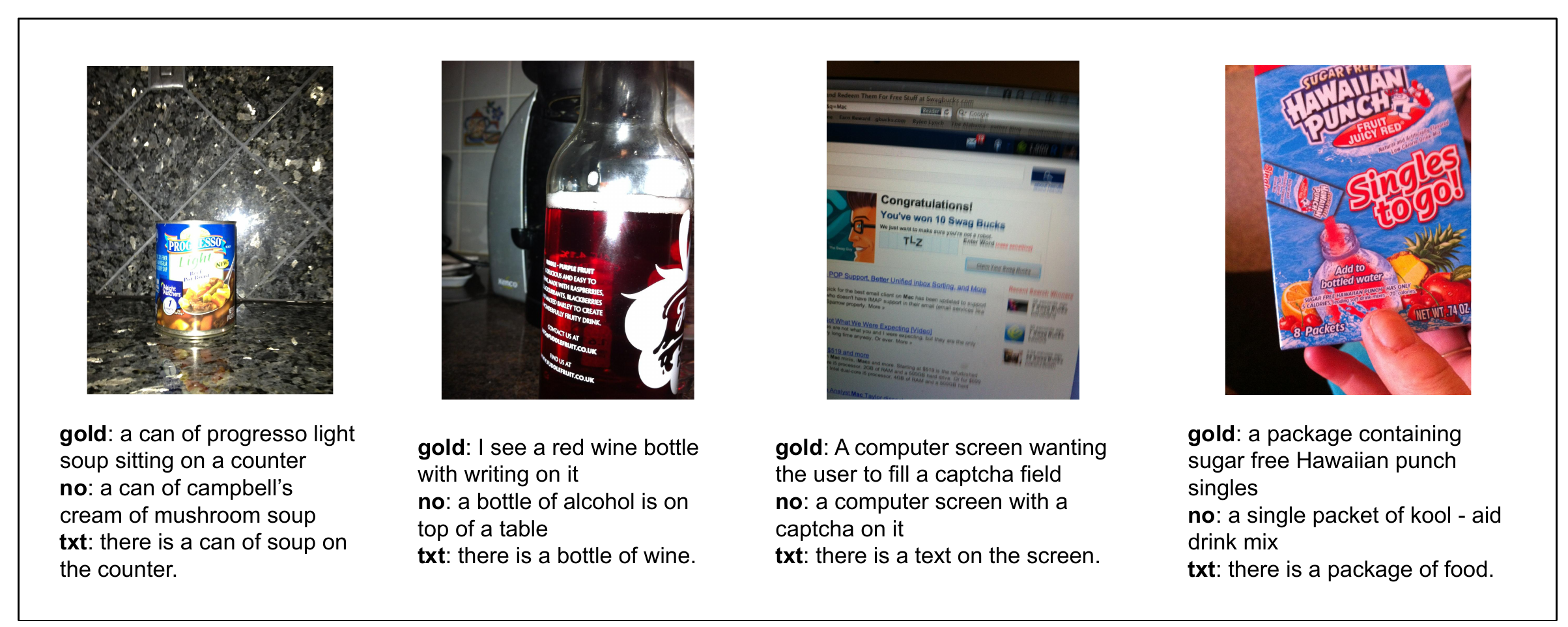}
  \caption{Generated captions without DA and with \textsc{txt} DA, compared with one of the gold captions.}
  \label{fig:quality}
\end{figure*}

VizWiz contains some images of bad quality: in some cases, the caption reads \textit{'Quality issues are too severe to recognize visual content'}. In order to avoid the generation of these captions during inference, they can be removed from the training set \cite{9909624}. In our work, we exclude an image from training, if at least 3 out of the five captions in the image contain this caption; that means that more than 50\% of the annotators could not describe the content of the image. If \textit{Quality Issues} are brought up only once or twice, we remove this caption and duplicate one or two of the other captions, so that, in the end, each image is annotated with five captions. We do not remove \textit{Quality Issues} images and captions from our test set. We exclude a total of 2,146 images.

While we technically do not use the complete dataset provided, it is justified by the fact that we test our pipeline in a low-resource scenario. \autoref{tab:stat} includes statistics over our tasks, including word type counts.

\section{Evaluation \& Results}\label{sec:results}

In this section, we present the evaluation metrics we used, our procedure, as well as the results from our core experiments. 

\subsection{Evaluation metrics \& splits}

Since IC is a natural language generation task, results are evaluated using standard metrics for evaluating text generation tasks. These metrics measure similarity to the ground truths. The metrics most commonly used are BLEU \cite{papineni-etal-2002-bleu}, ROUGE \cite{lin-2004-rouge}, METEOR \cite{banerjee-lavie-2005-meteor}, CIDEr \cite{vedantam-2015-cider}, and SPICE \cite{anderson2016spice}. For our hyperparameter tuning on the validation set, we use the BLEU metric. We report CIDEr scores in the main paper for brevity, scores for the other evaluation metrics can be found in \autoref{app:sec:results}. We use the Pycocoevalcap\footnote{\url{https://github.com/salaniz/pycocoevalcap.git}} library for evaluation. In order to evaluate the continual learning abilities of our IC model, we report scores per cluster, as well as micro-averages over the clusters trained so far.

\subsection{Results}

% General observations -- DA
We present our results in \autoref{tab:results2}. The use of our DA methods does not improve the results. Especially when \textsc{img} DA is involved, performance drops dramatically compared to the \textsc{no} DA baseline. This leads us to the conclusion that the DA operations we applied to the images were not suitable. Unexpectedly, we observe that \textsc{txt} DA does not improve results compared to the \textsc{no} DA baseline, which is in contrast to findings of previous work showing that caption augmentation is beneficial for low-resource IC \cite{atliha2020text}. 
We analyze this in more detail in \autoref{sec:analysis}.

\begin{table}%[]
\centering
\begin{tabular}{lrrrr}
\toprule
 & \textsc{no} & \textsc{img} & \textsc{txt} & \textsc{both} \\
 \midrule
no. of types & 1,383 & 2,418 & 1,397 & 1,053 \\
$\diameter$ (median) & 10.0 & 10.0 & 8.0 & 10.0 \\
$\diameter$ (mean) & 10.229 & 10.464 & 7.949 & 9.894 \\
\bottomrule
\end{tabular}
\caption{Statistics over captions generated with our models. $\diameter$ : average caption length.}
\label{tab:capstats}
\end{table}

\begin{table*}%[]
\centering
\resizebox{\linewidth}{!}{
\begin{tabular}{l|rrrr|rrrr|rrrr|rrrr|rrrr}
\toprule
 & \multicolumn{4}{c}{+ cluster 1} & \multicolumn{4}{c}{+ cluster 2} & \multicolumn{4}{c}{+ cluster 3} & \multicolumn{4}{c}{+ cluster 4} & \multicolumn{4}{c}{+ cluster 5} \\
 \midrule
DA & \multicolumn{2}{c}{\textsc{no}} & \multicolumn{2}{c}{\textsc{txt}} & \multicolumn{2}{c}{\textsc{no}} & \multicolumn{2}{c}{\textsc{txt}} & \multicolumn{2}{c}{\textsc{no}} & \multicolumn{2}{c}{\textsc{txt}} & \multicolumn{2}{c}{\textsc{no}} & \multicolumn{2}{c}{\textsc{txt}} & \multicolumn{2}{c}{\textsc{no}} & \multicolumn{2}{c}{\textsc{txt}} \\
\midrule
MEM & \multicolumn{1}{c}{+} & \multicolumn{1}{c}{-} & \multicolumn{1}{c}{+} & \multicolumn{1}{c}{-} & \multicolumn{1}{c}{+} & \multicolumn{1}{c}{-} & \multicolumn{1}{c}{+} & \multicolumn{1}{c}{-} & \multicolumn{1}{c}{+} & \multicolumn{1}{c}{-} & \multicolumn{1}{c}{+} & \multicolumn{1}{c}{-} & \multicolumn{1}{c}{+} & \multicolumn{1}{c}{-} & \multicolumn{1}{c}{+} & \multicolumn{1}{c}{-} & \multicolumn{1}{c}{+} & \multicolumn{1}{c}{-} & \multicolumn{1}{c}{+} & \multicolumn{1}{c}{-} \\
\midrule
1 & 27.1 & 27.1 & 20.9 & 20.8 & 16.5 & 15.6 & 14.3 & 9.7 & 22.8 & 22.9 & 17.7 & 16.3 & 19.3 & 20.8 & 13.1 & 15.2 & 17.6 & 19.0 & 13.3 & 13.5 \\
2 & & & & & 26.0 & 27.0 & 22.2 & 20.1 & 25.2 & 24.9 & 18.7 & 17.2 & 19.3 & 17.2 & 16.0 & 15.3 & 23.3 & 23.0 & 18.3 & 15.2 \\
3 & & & & & & & & & 32.4 & 31.3 & 28.1 & 24.1 & 24.2 & 23.8 & 17.9 & 18.4 & 25.1 & 24.2 & 18.1 & 19.1 \\
4 & & & & & & & & & & & & & 25.3 & 23.7 & 17.5 & 20.9 & 18.5 & 18.9 & 13.5 & 12.2 \\
5 & & &  &  & &  &  & & &  & &  &  &  &  &  & 27.1 & 25.6 & 19.9 & 19.0 \\
\midrule
all & 27.1 & 27.1 & 20.9 & 20.8 & \textbf{21.0} & 20.4 & \textbf{18.5} & 14.3 & \textbf{29.7} & 29.1 & \textbf{24.8} & 22.0 & 23.4 & \textbf{\textcolor{red}{23.5}} & 17.2 & \textbf{\textcolor{red}{18.1}} & \textbf{24.9} & 24.3 & \textbf{18.6} & 18.4 \\
\bottomrule
\end{tabular}
}
\caption{CIDEr results on the validation set for \textsc{no} and \textsc{txt} augmentation with (+) and without (-) episodic memory replay. We mark in \textbf{bold} the cases in which episodic memory contributes to an improvement, and in \textbf{\textcolor{red}{red}} the cases in which it does not.}
\label{tab:memplusminus}
\end{table*}

\begin{figure*}[ht]
\centering
  \includegraphics[width=\linewidth]{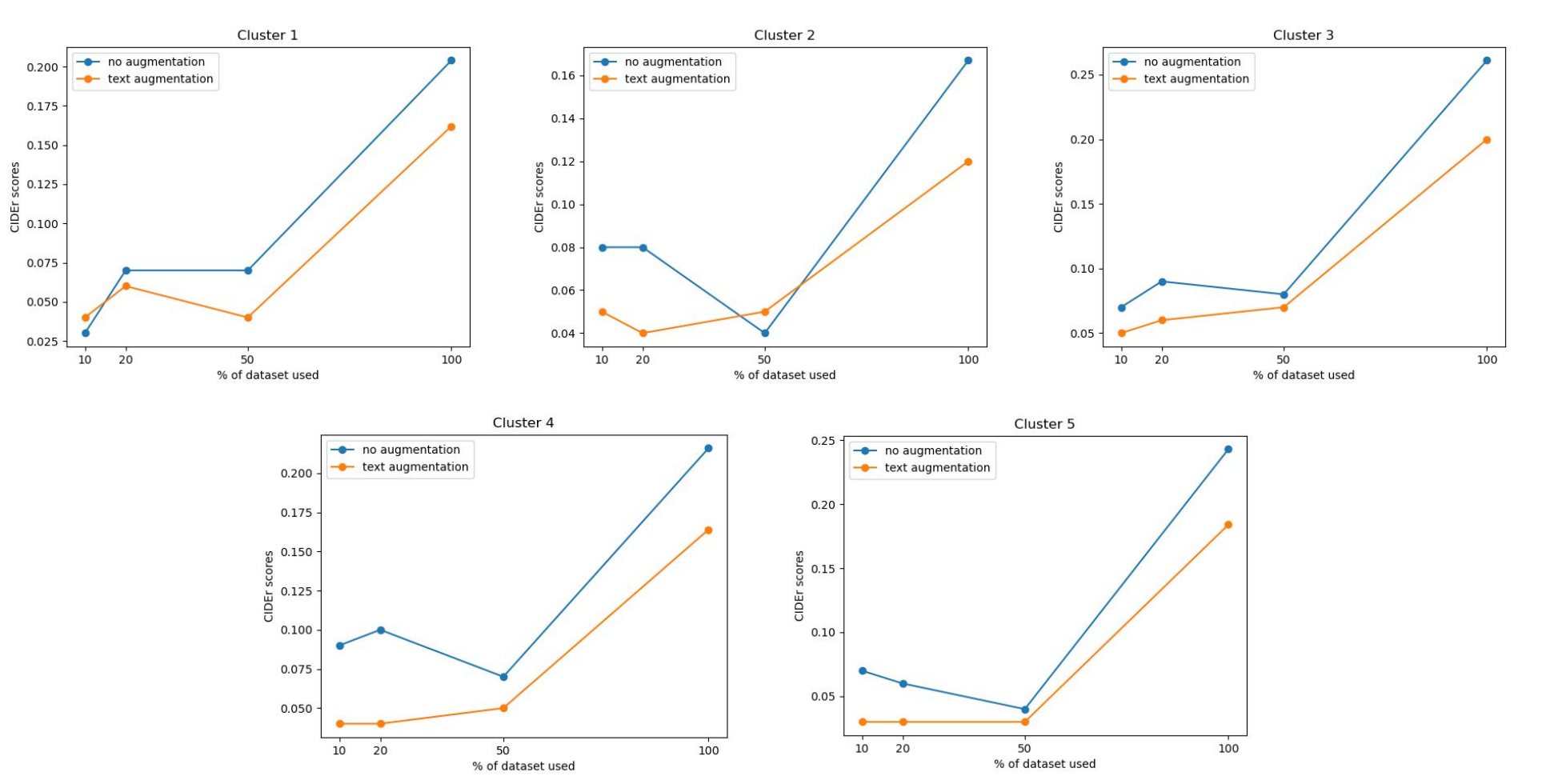}
  \caption{CIDEr results on the validation set for each task training with 10\%, 20\%, 50\%, and 100\% of the data.}
\label{fig:percent}
\end{figure*}

\section{Analysis}\label{sec:analysis}

In this section, we take a closer look into the quality of the captions generated by our models. We focus on the \textsc{no} and \textsc{txt} models since they produce better results. We also conduct two ablation studies: one considers training without the use of the memory module, and the other one tests our method in a low-resource scenario.

\subsection{Caption quality}

In order to gain a better insight into our results, in particular the observation that \textsc{txt} DA worsens results compared to the \textsc{no} DA baseline, we compare the generated captions based on their average length and the number of unique word types contained in the captions. One aspect that strikes immediately when comparing captions generated with \textsc{txt} DA vs \textsc{no} DA is variation. While we find that \textsc{no} captions and \textsc{txt} captions share a similar amount of unique word types, their average length is different, with \textsc{txt} captions being more than 2 words shorter than \textsc{no} captions.

We include some examples of generated captions in \autoref{fig:quality}. While we see that the captions generated are not necessarily erroneous, captions generated with the models trained with \textsc{txt} DA are less informative than the gold captions and captions generated without DA. 
Automated evaluation metrics often penalize changes in the length of the output. Captions generated by the \textsc{txt} DA model tend to be more similar to the paraphrases generated by the PEGASUS paraphrasing model (which was used to generate data for the training of the \textsc{txt} DA model), %used to train the \txtsc{txt} DA model,
which are shorter and less informative. Hence, this paraphrasing tool is not suitable for this particular task. In the future, we plan to compare more paraphrasing tools for DA on IC tasks.

To confirm our qualitative observations in a quantitative manner, we carried out a small manual analysis. We randomly sampled 100 captions generated with the \textsc{txt} models and compared them to the gold captions. Our criterion was informativeness: we ranked each generated caption as \textit{non-informative}, \textit{partially informative}, or \textit{very informative}. We find that 46 of them are very or partially informative, while for some of the rest, the lack of informativeness comes from the fact that the image quality is low (since seven of them contain severe quality issues). % We conducted the annotation ourselves: need to clear this up!

\subsection{Ablation study: Training without episodic memory replay}

In order to investigate the effect of the sparse episodic memory replay on the continual learning abilities of the model, we train models in the same settings as in our core experiments, except for the use of sparse episodic memory replay. Results for these experiments are shown in \autoref{tab:memplusminus}. We observe that, in general, there is an improvement in performance in almost all cases, both in models trained with \textsc{no} DA and in models trained with \textsc{txt} DA. The only exception is the model after training with cluster 4, which is significantly smaller than the rest of the other clusters (approx. 1/3 the size of the next smaller cluster). This shows that, while the episodic memory module positively influences performance when at least 1000 samples are present, it is not as effective with very few samples.

\subsection{Ablation study: Training with parts of the dataset} 

In an interactive setup, we cannot assume large amounts of annotated data provided by the user, hence we evaluate our models after training on only 10\%, 20\%, and 50\% of the data of each cluster. Training data points for each cluster are chosen randomly - for this reason, we present average scores over 3 trainings with the same settings. Our training takes place without memory since in most cases, the amount of data is too small for the memory to be activated. The results for models trained on reduced amounts of data for each cluster are shown in \autoref{fig:percent}.

It seems that \textsc{txt} DA does not improve results even in a low-resource scenario - the curves for \textsc{no} and \textsc{txt} DA are similar for the larger clusters (1, 3, 5). For task 2, a slight improvement in performance can be observed when training with 50\% of the data. This, in turn, leads to an additional observation, namely the fact that almost all our \textsc{no} DA models deter when trained with half of the data of each cluster. This might be attributed to the data distribution of the clusters with which we trained.

\section{Conclusion}\label{sec:conclusion}

We have presented a pipeline for interactive IC, which combines simple methods for incremental model improvement. This framework allows incremental adaptation of a pre-trained IC model to new data that is entered by users. The user input is transformed into a larger data batch using various data augmentation methods (for image, text, and both modalities). We additionally adapted a continual learning method for IC, which prevents catastrophic forgetting after repeated updates. In order to simulate incremental user input, we split the relatively small, domain-specific VizWiz dataset into non-overlapping clusters based on nouns mentioned in the image captions. VizWiz is a good test bed for our pipeline, as it contains real-world images with varying quality. 

We analyzed the effectiveness of DA in our experiments, and we noticed a lower performance of our models when trained with augmented data. The drop in performance resulting from the application of DA methods was evident in our low-resource experiments as well. We concluded that, especially for IC, \textsc{img} DA must be applied carefully. The same applies to \textsc{txt} DA: since brevity is penalized in this task, the DA outputs should be of similar length and descriptiveness as the gold captions. We confirmed that sparse memory replay does enable the models to retain knowledge learned from previous datasets while adapting to new data.

In the future, we plan to experiment with more elaborate joint DA methods for IC. Apart from evaluating the approach with respect to model performance using automated performance metrics, we intend to evaluate its usefulness and usability for end-users in a human study. Since prompting using large models is a popular paradigm recently, we intend to experiment with models like CLIP \cite{radford2021clip} as well, additionally assessing the trade-off between initial training cost and adaptation cost. Last but not least, applying active learning methods to select the best sample(s) for the episodic memory module can potentially increase the effectiveness of the continual learning method used in our pipeline.

\section*{Limitations}

Despite the promising results of our IML pipeline for image captioning, our work has some limitations. Firstly, the experiments were conducted on a domain-specific dataset, VizWiz, and may not generalize to other datasets or domains. Secondly, our approach may not be suitable for scenarios where user feedback is sparse or unreliable, as the effectiveness of IML heavily depends on the quality and quantity of the feedback. Thirdly, our use of episodic memory to retain knowledge from previously seen clusters may not scale well to smaller datasets and other methods may be required. Lastly, our approach does not address the challenge of bias in the data, which can lead to biased models. %and results.

\section*{Ethical Statement}

As of now, we do not see ethical concerns with the study presented in this paper. We used a dataset that is publicly available. The study is currently not applied to human subjects with personal data; in this case, the use of user feedback in the training process could potentially introduce biases if the feedback is not diverse or representative of the population. Lastly, our approach may be used to develop image captioning models that generate harmful or inappropriate content, such as captions that perpetuate harmful stereotypes or stigmatize certain groups of people. % We must ensure that our models are developed and used responsibly, with appropriate safeguards in place to prevent harm.

\section*{Acknowledgments}

We thank the reviewers for their insightful comments and suggestions. The research was funded by the XAINES project (BMBF, 01IW20005) and by the No-IDLE project (BMBF, 01IW23002).

% \section*{Acknowledgments}

% We thank the reviewers for their insightful comments and suggestions. The research was funded by the XAINES project (BMBF, 01IW20005).

% The preparation of these instructions and the \LaTeX{} and Bib\TeX{}
% files that implement them was supported by Schlumberger Palo Alto
% Research, AT\&T Bell Laboratories, and Morgan Kaufmann Publishers.
% Preparation of the Microsoft Word file was supported by IJCAI.  An
% early version of this document was created by Shirley Jowell and Peter
% F. Patel-Schneider.  It was subsequently modified by Jennifer
% Ballentine and Thomas Dean, Bernhard Nebel, Daniel Pagenstecher,
% Kurt Steinkraus, Toby Walsh and Carles Sierra. The current version
% has been prepared by Marc Pujol-Gonzalez and Francisco Cruz-Mencia.

% \appendix

\bibliographystyle{acl_natbib}
\bibliography{ijcai23}

% \clearpage
% \newpage

\appendix

\section{Example NPs for VizWiz clustering}\label{app:sec:nps}

We include example nominal phrases (NPs) from our VizWiz clustering. We follow the procedure described in the main body of the paper. For each cluster, we include 20 NPs. While there is no perfect separation in object categories, we do notice certain semantic similarities between the NPs in most clusters:

% \begin{itemize}
%     \item Cluster 1: Technology
%     \item Cluster 2: Brand names
%     \item Cluster 3: Miscellaneous
%     \item Cluster 4: Food-related vocabulary
%     \item Cluster 5: Furniture
% \end{itemize}

\begin{table*}[bp]
\centering
\resizebox{\linewidth}{!}{
\begin{tabular}{c|c|c|c|c}
\toprule
cluster 1 & cluster 2 & cluster 3 & cluster 4 & cluster 5 \\ \midrule
gift card & ac & kitchen   counter top & ingredients & dark   surface \\
button & labrador & top   portion & small   packet & glass   cup \\
camera lens & quaker & small   dog & crock   pot & light   fixture \\
nutrition information & stouffer & bottom & large   bottle & wooden   countertop \\
apple & dr & left side & nutritional & beige   carpet \\
video games & packet & small & kitchen   appliance & black \\
electrical outlet & screenshot & eye   drops & ingredients   label & lamp \\
tv screen & sainsbury & math   problems & lotion   bottle & wire \\
cable box & barcode & paper   money & milk   chocolate & concrete   floor \\
computer tower & coke & led & liter   bottle & interior \\
tv & nokia & person   's knee & dark   chocolate & plastic   container \\
cd case & samsung & 's   chicken & medicine   bottle & marble   counter \\
silver device & tan & brand   name & frozen   dinner box & glass   container \\
keys & unopened & side   view & dinner   table & shorts \\
image quality & container/   box / bottle & counter top & water   bottle & styrofoam \\
design & sprite & sunny   day & small   jar & couch   cushion \\
entertainment center & the/this & remote   control & spice & plastic   wrapping \\
book page & roni & body & coffee   pod & glass   door \\
background & k-cup & room   area & brownie   mix & clear   plastic bag \\
laptop monitor & upc & left   side & ice   cream & flat   horizontal surface \\
\bottomrule
\end{tabular}
}
\caption{First 20 NPs for each cluster from the VizWiz Dataset}
\label{app:vizwiz}
\end{table*}

\section{Results for BLEU-4, METEOR, ROUGE, SPICE metrics}\label{app:sec:results}

In the main paper, we only include CIDEr scores for our main experiments. Here we present results in four additional metrics: BLEU-4 (\autoref{app:bleu}), METEOR (\autoref{app:meteor}), ROUGE-L (\autoref{app:rouge}), and SPICE (\autoref{app:spice}). The tables can be found on the next page.

\begin{table*}[]
\centering
\resizebox{\linewidth}{!}{
\begin{tabular}{l|rrrr|rrrr|rrrr|rrrr|rrrr}
\toprule
 & \multicolumn{4}{c}{+ cluster 1} & \multicolumn{4}{c}{+ cluster 2} & \multicolumn{4}{c}{+ cluster 3} & \multicolumn{4}{c}{+ cluster 4} & \multicolumn{4}{c}{+ cluster 5} \\
 \midrule
DA & \multicolumn{1}{c}{\textsc{no}} & \multicolumn{1}{c}{\textsc{img}} & \multicolumn{1}{c}{\textsc{txt}} & \multicolumn{1}{c}{\textsc{both}} & \multicolumn{1}{c}{\textsc{no}} & \multicolumn{1}{c}{\textsc{img}} & \multicolumn{1}{c}{\textsc{txt}} & \multicolumn{1}{c}{\textsc{both}} & \multicolumn{1}{c}{\textsc{no}} & \multicolumn{1}{c}{\textsc{img}} & \multicolumn{1}{c}{\textsc{txt}} & \multicolumn{1}{c}{\textsc{both}} & \multicolumn{1}{c}{\textsc{no}} & \multicolumn{1}{c}{\textsc{img}} & \multicolumn{1}{c}{\textsc{txt}} & \multicolumn{1}{c}{\textsc{both}} & \multicolumn{1}{c}{\textsc{no}} & \multicolumn{1}{c}{\textsc{img}} & \multicolumn{1}{c}{\textsc{txt}} & \multicolumn{1}{c}{\textsc{both}} \\
\midrule
eval on 1 & 14.4 & 6.0 & 11.1 & 12.4 & 9.1 & 2.6 & 8.7 & 4.6 & 11.6 & 2.8 & 9.2 & 6.3 & 10.5 & 2.6 & 7.5 & 4.5 & 7.6 & 2.9 & 6.0 & 6.1 \\
eval on 2 & & & & & 16.9 & 7.4 & 15.1 & 13.7 & 17.8 & 5.6 & 15.9 & 10.6 & 16.6 & 5.7 & 12.5 & 9.7 & 16.2 & 4.8 & 12.5 & 13.6 \\
eval on 3 & & & & & & & & & 16.0 & 4.9 & 13.8 & 11.3 & 13.7 & 3.5 & 10.5 & 6.4 & 13.8 & 3.8 & 10.8 & 10.6 \\
eval on 4 & & & & & & & & & & & & & 16.9 & 4.6 & 13.5 & 9.1 & 12.4 & 3.3 & 9.8 & 8.4 \\
eval on 5 & & & & & & & & & & & & & & & & & 15.1 & 4.4 & 11.3 & 12.1 \\
\midrule
micro avg & 14.4 & 6.0 & 11.1 & 12.4 & 10.4 & 3.5 & 9.8 & 6.3 & 14.0 & 4.0 & 11.8 & 8.9 & 12.5 & 3.4 & 9.4 & 6.0 & 12.4 & 3.8 & 9.6 & 9.9 \\
\bottomrule
\end{tabular}
}
\caption{BLEU-4 results on our experiments on VizWiz data clustered according to the procedure described in our main paper. We start with the model resulting from the supervised pre-training step on MS COCO and continue to train this model incrementally on the VizWiz clusters (+cluster ...). We include the amount of (original) training data in brackets. DA: Data augmentation, \textsc{no}: no DA, \textsc{img}: image DA, \textsc{txt}: text DA, \textsc{both}: image and text DA. The numbers in the left column stand for clusters evaluated on. 'all' refers to the micro average score.}
\label{app:bleu}
\end{table*}

\begin{table*}[]
\centering
\resizebox{\linewidth}{!}{
\begin{tabular}{l|rrrr|rrrr|rrrr|rrrr|rrrr}
\toprule
 & \multicolumn{4}{c}{+ cluster 1} & \multicolumn{4}{c}{+ cluster 2} & \multicolumn{4}{c}{+ cluster 3} & \multicolumn{4}{c}{+ cluster 4} & \multicolumn{4}{c}{+ cluster 5} \\
 \midrule
DA & \multicolumn{1}{c}{\textsc{no}} & \multicolumn{1}{c}{\textsc{img}} & \multicolumn{1}{c}{\textsc{txt}} & \multicolumn{1}{c}{\textsc{both}} & \multicolumn{1}{c}{\textsc{no}} & \multicolumn{1}{c}{\textsc{img}} & \multicolumn{1}{c}{\textsc{txt}} & \multicolumn{1}{c}{\textsc{both}} & \multicolumn{1}{c}{\textsc{no}} & \multicolumn{1}{c}{\textsc{img}} & \multicolumn{1}{c}{\textsc{txt}} & \multicolumn{1}{c}{\textsc{both}} & \multicolumn{1}{c}{\textsc{no}} & \multicolumn{1}{c}{\textsc{img}} & \multicolumn{1}{c}{\textsc{txt}} & \multicolumn{1}{c}{\textsc{both}} & \multicolumn{1}{c}{\textsc{no}} & \multicolumn{1}{c}{\textsc{img}} & \multicolumn{1}{c}{\textsc{txt}} & \multicolumn{1}{c}{\textsc{both}} \\
\midrule 
eval on 1 & 13.5 & 9.3 & 12.4 & 12.9 & 10.8 & 6.7 & 10.5 & 7.8 & 12.4 & 6.8 & 10.7 & 8.8 & 11.2 & 6.5 & 9.7 & 7.7 & 10.8 & 7.3 & 9.4 & 9.3 \\
eval on 2 & & & & & 15.8 & 10.3 & 15.0 & 13.7 & 16.0 & 9.3 & 14.6 & 11.9 & 14.8 & 9.4 & 13.3 & 11.2 & 15.8 & 8.9 & 13.7 & 13.8 \\
eval on 3 & & & & & & & & & 15.2 & 8.6 & 13.7 & 12.0 & 13.9 & 7.7 & 12.0 & 9.4 & 14.1 & 8.5 & 12.2 & 11.9 \\
eval on 4 & & & & & & & & & & & & & 15.1 & 9.0 & 13.0 & 11.5 & 13.8 & 8.0 & 12.0 & 11.4 \\
eval on 5 & & & & & & & & & & & & & & & & & 15.4 & 9.3 & 13.3 & 13.0 \\
\midrule
micro avg & 13.5 & 9.3 & 12.4 & 12.9 & 11.6 & 7.3 & 11.2 & 8.7 & 13.9 & 7.8 & 12.3 & 10.5 & 12.8 & 7.3 & 11.0 & 8.8 & 13.6 & 8.4 & 11.7 & 11.5 \\
\bottomrule
\end{tabular}
}
\caption{METEOR results, as above.}\label{app:meteor}
\end{table*}

\begin{table*}[]
\centering
\resizebox{\linewidth}{!}{
\begin{tabular}{l|rrrr|rrrr|rrrr|rrrr|rrrr}
\toprule
 & \multicolumn{4}{c}{+ cluster 1} & \multicolumn{4}{c}{+ cluster 2} & \multicolumn{4}{c}{+ cluster 3} & \multicolumn{4}{c}{+ cluster 4} & \multicolumn{4}{c}{+ cluster 5} \\
 \midrule
DA & \multicolumn{1}{c}{\textsc{no}} & \multicolumn{1}{c}{\textsc{img}} & \multicolumn{1}{c}{\textsc{txt}} & \multicolumn{1}{c}{\textsc{both}} & \multicolumn{1}{c}{\textsc{no}} & \multicolumn{1}{c}{\textsc{img}} & \multicolumn{1}{c}{\textsc{txt}} & \multicolumn{1}{c}{\textsc{both}} & \multicolumn{1}{c}{\textsc{no}} & \multicolumn{1}{c}{\textsc{img}} & \multicolumn{1}{c}{\textsc{txt}} & \multicolumn{1}{c}{\textsc{both}} & \multicolumn{1}{c}{\textsc{no}} & \multicolumn{1}{c}{\textsc{img}} & \multicolumn{1}{c}{\textsc{txt}} & \multicolumn{1}{c}{\textsc{both}} & \multicolumn{1}{c}{\textsc{no}} & \multicolumn{1}{c}{\textsc{img}} & \multicolumn{1}{c}{\textsc{txt}} & \multicolumn{1}{c}{\textsc{both}} \\
\midrule 
eval on 1 & 34.0 & 26.0 & 31.0 & 31.5 & 28.3 & 21.0 & 28.8 & 23.1 & 31.3 & 22.2 & 29.0 & 26.9 & 30.8 & 21.6 & 26.9 & 24.8 & 29.9 & 23.6 & 27.0 & 28.4 \\
eval on 2 & & & & & 39.0 & 30.8 & 39.1 & 36.1 & 42.4 & 29.3 & 39.5 & 34.9 & 39.7 & 29.7 & 35.8 & 33.6 & 42.3 & 29.8 & 37.2 & 39.9 \\
eval on 3 & & & & & & & & & 39.8 & 27.2 & 37.2 & 34.3 & 37.4 & 25.6 & 33.2 & 29.8 & 38.7 & 27.8 & 34.5 & 35.3 \\
eval on 4 & & & & & & & & & & &  & & 38.0 & 27.1 & 34.4 & 32.2 & 36.4 & 26.8 & 34.0 & 33.7 \\
eval on 5 & & & & & & & & & & & & & & & & & 40.7 & 29.3 & 35.7 & 37.5 \\
\midrule
micro avg & 34.0 & 26.0 & 31.0 & 31.5 & 30.1 & 22.6 & 30.5 & 25.3 & 35.9 & 25.0 & 33.5 & 30.8 & 34.5 & 24.1 & 30.5 & 27.8 & 36.7 & 27.0 & 32.6 & 34.1 \\
\bottomrule
\end{tabular}
}
\caption{ROUGE-L results, as above.}\label{app:rouge}
\end{table*}

\begin{table*}[]
\centering
\resizebox{\linewidth}{!}{
\begin{tabular}{l|rrrr|rrrr|rrrr|rrrr|rrrr}
\toprule
 & \multicolumn{4}{c}{+ cluster 1} & \multicolumn{4}{c}{+ cluster 2} & \multicolumn{4}{c}{+ cluster 3} & \multicolumn{4}{c}{+ cluster 4} & \multicolumn{4}{c}{+ cluster 5} \\
 \midrule
DA & \multicolumn{1}{c}{\textsc{no}} & \multicolumn{1}{c}{\textsc{img}} & \multicolumn{1}{c}{\textsc{txt}} & \multicolumn{1}{c}{\textsc{both}} & \multicolumn{1}{c}{\textsc{no}} & \multicolumn{1}{c}{\textsc{img}} & \multicolumn{1}{c}{\textsc{txt}} & \multicolumn{1}{c}{\textsc{both}} & \multicolumn{1}{c}{\textsc{no}} & \multicolumn{1}{c}{\textsc{img}} & \multicolumn{1}{c}{\textsc{txt}} & \multicolumn{1}{c}{\textsc{both}} & \multicolumn{1}{c}{\textsc{no}} & \multicolumn{1}{c}{\textsc{img}} & \multicolumn{1}{c}{\textsc{txt}} & \multicolumn{1}{c}{\textsc{both}} & \multicolumn{1}{c}{\textsc{no}} & \multicolumn{1}{c}{\textsc{img}} & \multicolumn{1}{c}{\textsc{txt}} & \multicolumn{1}{c}{\textsc{both}} \\
\midrule 
eval on 1 & 7.5 & 5.3 & 7.2 & 7.2 & 5.0 & 1.7 & 5.0 & 2.1 & 6.2 & 1.5 & 5.1 & 3.1 & 5.0 & 1.5 & 4.6 & 1.8 & 4.6 & 1.5 & 4.0 & 3.3 \\
eval on 2 & & & & & 9.6 & 3.8 & 8.8 & 6.6 & 8.8 & 2.5 & 8.2 & 5.1 & 7.9 & 3.2 & 7.4 & 4.5 & 8.6 & 2.2 & 7.6 & 6.6 \\
eval on 3 & & & & & & & & & 8.3 & 2.4 & 7.3 & 5.6 & 6.8 & 1.7 & 5.7 & 2.7 & 7.0 & 1.9 & 5.8 & 4.8 \\
eval on 4 & & & & & & & & & & & & & 8.0 & 2.9 & 7.1 & 4.5 & 7.0 & 1.7 & 6.1 & 5.3 \\
eval on 5 & & & & & & & & & & & & & & & & & 8.5 & 2.6 & 7.5 & 6.2 \\
\midrule
micro avg & 7.5 & 5.3 & 7.2 & 7.2 & 5.8 & 2.0 & 5.6 & 2.9 & 7.3 & 1.9 & 6.3 & 4.3 & 6.1 & 1.8 & 5.4 & 2.5 & 6.8 & 2.0 & 5.9 & 4.9 \\
\bottomrule
\end{tabular}
}
\caption{SPICE results, as above.}\label{app:spice}
\end{table*}

\end{document}